\documentclass{article}

    \usepackage[preprint]{neurips_2019}

\usepackage[utf8]{inputenc} % allow utf-8 input
\usepackage[T1]{fontenc}    % use 8-bit T1 fonts
\usepackage{hyperref}       % hyperlinks
\usepackage{url}            % simple URL typesetting
\usepackage{booktabs}       % professional-quality tables
\usepackage{amsfonts}       % blackboard math symbols
\usepackage{nicefrac}       % compact symbols for 1/2, etc.
\usepackage{microtype}      % microtypography

\usepackage[square,numbers]{natbib}

\usepackage{microtype}
\usepackage{graphicx}
\usepackage{subfigure}
\usepackage{booktabs} % for professional tables
\usepackage{times}
\usepackage{helvet}
\usepackage{courier}
\usepackage{amsmath}
\usepackage{bm}
\usepackage{dsfont}
\usepackage{paralist}
\usepackage{hyperref}
\input{taesup.sty}

\newcommand{\eg}{{\it e.g.}}
\newcommand{\ie}{{\it i.e.}}

\newcommand{\Pib}{\mathbf{\Pi}}
\newcommand{\Lb}{\mathbf{\Lambda}}
\newcommand{\Ellb}{\mathbf{L}}

\newcommand{\lambdab}{\boldsymbol{\lambda}}

\title{Iterative Channel Estimation for\\ Discrete Denoising under Channel Uncertainty}

\author{%
  Hongjoon Ahn, Taesup Moon \\
  Department of Electrical and Computer Engineering\\
  Sungkyunkwan University\\
  Suwon, Korea 16419 \\
  \texttt{\{hong0805,tsmoon\}@skku.edu} \\
}

\begin{document}

\maketitle

\begin{abstract}
We propose a novel iterative channel estimation (ICE) algorithm that essentially removes the critical known noisy channel assumption for universal discrete denoising problem. Our algorithm is based on Neural DUDE (N-DUDE), a recently proposed neural network-based discrete denoiser, and it estimates the channel transition matrix as well as the neural network parameters in an alternating manner until convergence. While we do not make any probabilistic assumption on the underlying clean data, our ICE resembles Expectation-Maximization (EM) with variational approximation, and it takes advantage of the property that N-DUDE can always induce a marginal posterior distribution of the clean data. We carefully validate the channel estimation quality of ICE, and with extensive experiments on several radically different types of data, we show the ICE equipped neural network-based denoisers can perform \emph{universally} well regardless of the uncertainties in both the channel and the clean source. Moreover, we show ICE becomes extremely robust to its hyperparameters, and show our denoisers with ICE significantly outperform the strong baseline that can handle the channel uncertainties for denoising, the widely used Baum-Welch (BW) algorithm for hidden Markov models (HMM).
\end{abstract}

%!TEX root = ch_est_aaai.tex

\section{Introduction}
\label{introduction}

Denoising, which focuses on cleaning up noise-corrupted data, is one of the most studied topics in machine learning and signal processing. In particular, \emph{discrete denoising} focuses on denoising the data that take finite-alphabet values. Such setting covers several applications in various domains, \eg, image denoising  \cite{Ordetal03,MotOrdRamSerWei11}, DNA sequence denoising \cite{LaeBorMcH16,LeeMooYooWei16}, and channel decoding \cite{OrdSerVerVis08}, etc. Recently, utilizing quantized measurements from low-power sensors \cite{Rom17} or DNA sequencing devices \cite{nanocorr} are getting more prevalent, thus, denoising such data is getting more important.

In \cite{Dude}, the \emph{universal} setting for discrete denoising, in which no assumption on the underlying clean data was made, was first considerd. They devised a sliding-window algorithm called DUDE (Discrete Universal DEnoiser) with powerful theoretical guarantees and empirical performance. Despite the strong results, however, DUDE suffered from a couple of shortcomings; the performance of the algorithm deteriorates as the alphabet size grows and is quite sensitive to the choice of a hyperparameter, the window size $k$.
% , and there was no specific way of choosing $k$ for the given noisy data. 
In order to overcome such limitations, \cite{MooMinLeeYoo16} recently proposed Neural DUDE (N-DUDE), by introducing a neural network as an implicit context aggregator. It maintained the robustness with respect to both $k$ and the alphaset size, and as a result, N-DUDE achieved significantly better performance than DUDE. The main gist of N-DUDE was to devise ``pseudo-labels'' solely based on the noisy data, and train the neural network as a denoiser without any supervised training set with clean data. 
% More recently, as a follow-up of N-DUDE, CUDE (Context-aggregated Universal DEnoiser) was also proposed and achieved a better performance than N-DUDE.  

% As a result, the experimental result of Neural DUDE was very strong and significantly outperformed DUDE. 

Although both DUDE and N-DUDE did not make any assumptions on the underlying clean data, one critical assumption they both made is that the statistical characteristics of the noise mechanism is \emph{known} to the denoiser. That is, the noise is modeled to be a Discrete Memoryless Channel (DMC), \ie, the index-independent noise, and the channel transition matrix was assumed to be completely known to the denoiser. While such assumption makes sense in some applications, \eg, when the noisy channel can be reliably estimated with known reference sequences, it can become a major weakness in competing with other methods that do not require such assumption. For example, the Baum-Welch (BW) algorithm \cite{BaumPetrie70} combined with forward-backward (FB) recursion for hidden Markov models (HMM) \cite{EphMer02} can both estimate the channel (\ie, the emission probability) and the underlying clean data (\ie, the latent states) as long as the noisy observation can be modeled as an HMM.

In this paper, we aim to remove the known noise assumption of N-DUDE. Namely, the \emph{only} assumption we make is that the noise mechanism is a DMC (like in HMM), but neither the channel transition matrix nor characteristics of the clean data (such as Markovity) are assumed to be known. Thus, our setting is a much more challenging one than that of \cite{Dude,MooMinLeeYoo16} as we impose uncertainty on the noise model in addition to on the clean data\footnote{Such setting was initially considered in \cite{GemSigWes06,GemSigWes06a}, but mainly with a theoretical motivation.}. We propose a novel iterative channel estimation (ICE) algorithm such that learning the channel transition matrix and the neural network parameters can be done in an alternating manner that resembles Expectation-Maximization (EM). 
% The main motivation  our algorithm is from the observation that N-DUDE tends to be robust \emph{locally} around the true noisy channel model; thus, we need to estimate the channel up to a level that can attain the performance of the known channel counterpart of N-DUDE.  
The key component of our algorithm is to approximate the ``marginal posterior distribution'' of the clean data, even though there might not exist one, with a posterior induced from the N-DUDE's output distribution and carry out the variational approximation.

% rather than exactly estimating it. 
% and the objective of our learning is to minimize the denoising loss rather than accurate estimate of the channel. 

In our experimental results with various different types of data (\eg, images or DNA sequences), we show the effectiveness of our ICE by showing that denoising with the \emph{estimated} channel achieves almost the identical denoising performance as with the \emph{true} channel. We employ two neural network-based denoisers to evaluate the denoising performance with ICE; N-DUDE and CUDE (Context-aggregated Universal DEnoiser)\citep{RyuKim18}, in which the former is what ICE is based on, and the latter is another recently developed universal denoiser that is shown to outperform N-DUDE. Both algorithms that plug-in the estimated channel by ICE 
% (dubbed as ICE-N-DUDE and ICE-CUDE, respectively) 
are shown to outperform the widely used BW with FB recursion, which models the noisy data as an HMM regardless of it being true. In addition, we show ICE is much more robust with respect to its hyperparameters and initializations compared to BW, which is sensitive to the initial transition and channel models. Finally, we give thorough experimental analyses on the channel estimation errors as well as the model approximation performance of ICE.

\section{Notations and Related Work}\label{sec:notation_prelim}
% \subsection{Notations and problem setting}\label{subsec:notation_setting}

To be self-contained, we introduce notations that mainly follow \cite{MooMinLeeYoo16}. Throughout the paper, an $n$-tuple sequence is denoted as, \eg, $a^n=(a_1,\ldots,a_n)$, and $a_i^j$ refers to the subsequence $(a_i,\ldots,a_j)$. We denote the uppercase letters as random variables and the lowercase letters as either the realizations of the random variables or the individual symbols.
We denote $\Delta^d$ as the probability simplex in $\mathbb{R}^d$. 
In the \emph{universal} setting, the clean, underlying source data will be denoted as an \emph{individual sequence} $x^n$ as we make no stochastic assumption on it. We assume each component $x_i$ takes a value in some finite set $\mcX$. For example, for binary data, $\mcX=\{0,1\}$, and for DNA data, $\mcX=\{\texttt{A},\texttt{C},\texttt{G},\texttt{T}\}$. 
% Furthermore, as usual, the uppercase letters stand for the random variables, and the lowercase letters stand for the realizations of the random variables or the individual (non-random) symbols. 
% We denote the underlying source data as $\{X_i\}$ and assume each component takes values in some finite set $\mcX$. 

We assume $x^n$ is corrupted by a DMC, namely, the index-independent noise, and results in the noisy data, $Z^n$, of which each $Z_i$ takes a value in, again, a finite set $\mcZ$.
% The resulting noisy version of the source corrupted by a DMC is denoted as $\{Z_i\}$, and its components take values in, again, some finite set $\mcZ$. 
The DMC is characterized by the channel transition matrix $\mathbf\Pi\in\mathbb{R}^{|\mcX|\times|\mcZ|}$, and the $(x,z)$-th element of $\Pib$ stands for $\text{Pr}(Z=z|x)$. A natural assumption we make is that $\mathbf{\Pi}$ is of the \emph{full row rank}.
 We also denote $\Pib^\dagger=\Pib^\top(\Pib\Pib^\top)^{-1}$ as the Moore-Penrose pseudoinverse of $\Pib$. 
% In our setting, $\Pib$ is assumed to be known to the denoiser. 
% Furthermore, throughout this paper, we generally denote a sequence ($n$-tuple) as, \eg, $a^n=(a_1,\ldots,a_n)$, and $a_i^j$ refers to the subsequence $(a_i,\ldots,a_j)$.
% \begin{figure}[H]
%     \centering
%     \includegraphics[width=0.7\textwidth]{figures/background-setting.pdf}
%     \caption{The general setting of discrete denoising.}
%     \label{fig:general-setting}
% \end{figure}
% As shown in Fig.~\ref{fig:general-setting}, 
Now, upon observing the entire noisy data $Z^n$, a discrete denoiser reconstructs the original data with $\hat{X}^n=(\hat{X}_1(Z^n),\ldots,\hat{X}_n(Z^n))$, where each reconstructed symbol $\hat{X}_i(Z^n)$ takes its value in a finite set $\hat{\mathcal{X}}$. The goodness of the reconstruction is measured by the average denoising loss,
$
\frac{1}{n}\sum_{i=1}^n\Lb(x_i,\hat{X}_i(Z^n)),
% \label{eq:avg_loss}
$
where the per-symbol loss $\Lb(x_i,\hat{x}_i)$ measures the loss incurred by estimating $x_i$ with $\hat{x}_i$. The loss is fully represented with a loss matrix $\mathbf{\Lambda}\in\mathbb{R}^{|\mcX|\times|\hat{\mcX}|}$. 

% From now on, we will assume $|\mcX|=|\mcZ|=|\mcXhat|$ for simplicity. 

The $k$-th order sliding window denoisers are the denoisers that are defined by a time-invariant mapping $s_k:\mcZ^{2k+1}\rightarrow\hat{\mcX}$. That is, $\hat{X}_i(Z^n)= s_k(Z_{i-k}^{i+k})$. We also denote the tuple $(Z_{i-k}^{i-1},Z_{i+1}^{i+k})\triangleq\Cb_i$ as the $k$-th order double-sided context
% \footnote{Note we are using the uppercase notation $\Cb_i$ to highlight the randomness as opposed to $\cb_i$ used in \cite{MooMinLeeYoo16}.} 
around the noisy symbol $Z_i$, and we let $\mathbf{C}[k]$ as the set of all such contexts. 
% As discussed in \cite{MooMinLeeYoo16}, both DUDE in \cite{Dude} and Neural DUDE are sliding window denoisers.
We also denote $\mcS\triangleq\{s:\mcZ\rightarrow\hat{\mcX}\}$ as the set of \emph{single-symbol denoisers}  that are sliding window denoisers with $k=0$. Note $|\mcS|=|\hat{\mcX}|^{|\mcZ|}$. Then, an alternative view of of $s_k(\cdot)$ is that $s_k(\Cb_i,\cdot)\in\mcS$ is a single symbol denoiser defined by $\Cb_i$ and applied to $Z_i$. 

% Throughout the paper, for simplicity, we will assume $\mcX=\mcZ=\hat{\mcX}$, thus, assume that $\mathbf{\Pi}$ is invertible. 

% \subsection{Unbiased Estimated Loss}\label{subsec:main tool}
% \subsubsection{Main tool: Unbiased estimated loss}

% A critical building block of our ICE and Neural DUDE is the \emph{unbiased estimated} loss function as described in \citep[Section 3.1]{MooMinLeeYoo16}. That is, 

When $\Pib$ is known, as in \citep[Section 3.1]{MooMinLeeYoo16}, we can devise an \emph{unbiased esimate} of the true loss $\Lb$ as 
\be
\Ellb=\Pib^\dagger\bm{\rho}\in\mathbb{R}^{|\mcZ|\times|\mcS|},\label{eq:est_loss}
\ee in which $\bm{\rho}\in\mathbb{R}^{|\mcX|\times|\mcS|}$ with the $(x,s)$-th element is $\mathbb{E}_{Z|x}\Lb(x,s(Z))$, and $\mathbb{E}_{Z|x}(\cdot)$ stands for the expectation with respect to the distribution defined by the $x$-th row of $\Pib$. Then, as shown in \cite{MooMinLeeYoo16,UFP06}, $\Ellb$ has the unbiased property, $\mathbb{E}_{Z|x}\Ellb(Z,s)=\mathbb{E}_{Z|x}\Lb(x,s(Z))$.

\subsection{Related work}\label{subsec:related}

\noindent \textbf{DUDE \cite{Dude}} is a two-pass, sliding-window denoiser that has a linear complexity in data size $n$. For reconstruction at location $i$, DUDE takes $\Cb_i\in\Cb[k]$ and $Z_i\in\mcZ$, and applies the rule 
\begin{eqnarray}
\hat{X}_{i,\texttt{DUDE}}(\Cb_i,Z_i) = \arg\min_{\hat{x}\in\hat{\Xcal}}\hat{\mathbf{p}}_{\text{emp}}(\cdot|\Cb_i)^\top\Pib^{\dagger}[\boldsymbol\Lambda_{\hat{x}}\odot\Pib_{Z_i}],\label{eq:dude_rule}
\end{eqnarray}
in which $\hat{\mathbf{p}}_{\text{emp}}(\cdot|\Cb_i)$ is an empirical probability vector on $Z_i$ given the context vector $\Cb_i$, obtained from the noisy data $Z^n$. That is, for a context $\Cb\in\Cb[k]$, the $z$-th element becomes
\be
\hat{\mathbf{p}}_{\text{emp}}(z|\Cb)=\frac{|\{j:\Cb_j=\Cb, Z_j=z\}|}{|\{j:\Cb_j=\Cb\}|}.\label{eq:emp}
\ee
Moreover, the $\boldsymbol\Lambda_{\hat{x}}$ and $\Pib_{Z_i}$ in (\ref{eq:dude_rule}) stand for the $\hat{x}$-th and $Z_i$-th column of $\Lb$ and $\Pib$, respectively. The rule (\ref{eq:dude_rule}) is based on the intuition from achieving the Bayes response using (\ref{eq:emp}) and inverting the channel. For more rigorous details and results, we refer to the original paper \cite{Dude}.

% \cite{Dude} showed the \emph{universality} of DUDE in the sense that it can attain the performance of the best sliding-window denoiser for any underlying sequence $x^n$, provided that $k$ grows appropriately with $n$.
% for any underlying stationary process, it asymptotically attains the Bayes optimal preformance provided that $k$ grows appropriately with $n$. 
% For more details, we refer to the original paper \cite{Dude}.

% The main gist of \citep{MooMinLeeYoo16} was to devise the ``pseudo-labels'' based on the unbiased estimated loss matrix $\Ellb$ and use them as target labels when training a DNN-based sliding window denoiser. 

\noindent\textbf{Neural DUDE \cite{MooMinLeeYoo16}}
% takes the alternative view on the sliding-window denoiser mentioned above and 
extends DUDE and defines a \emph{single} neural network-based sliding-window denoiser, $\mathbf{p}^k(\mathbf{w},\cdot):\mathcal{Z}^{2k}\rightarrow\Delta^{|\mcS|}$, in which $\wb$ stands for the parameters in the network. Following the alternative view on the sliding-window denoiser mentioned above, the network
% It defines a neural network, $\mathbf{p}^k(\mathbf{w},\cdot):\mathcal{Z}^{2k}\rightarrow\Delta^{|\mcS|}$, 
takes the context $\Cb_i$ and outputs a probability distribution on the single-symbol denoisers to apply to $Z_i$.

To train $\wb$, N-DUDE defines the ``pseudo-label'' matrix $\Ellb_{\text{new}}\in\mathbb{R}^{|\mcZ|\times|\mcS|}$  as 
\be
\Ellb_{\text{new}} \triangleq -\Ellb+L_{\text{max}}\mathbf{1}_{|\mcZ|}\mathbf{1}_{|\mcS|}^\top,\label{eq:L_max}
\ee
in which $L_{\text{max}}\triangleq \max_{z,s}\Ellb(z,s)$, and $\mathbf{1}_{|\mcZ|}$ and $\mathbf{1}_{|\mcS|}$ stand for the all-1 vectors with $|\mcZ|$ and $|\mcS|$ dimensions, respectively. 
% Note $\Ellb_{\text{new}}$ can be computed solely based on $\Lb$ and the assumption of known $\mathbf{\Pi}$. 
By design, all the elements in $\Ellb_{\text{new}}$ are non-negative and can be computed with $\Pib$, $\Lb$, $z$ and $s$ (and \emph{not} with the clean $x$). 
% and $\mathbb{E}_x\Ellb_{\text{new}}(Z,s)=-\mathbb{E}_x\Lb(x,s(Z))+L_{\text{max}}$ from the unbiasedness of $\mathbf{L}$; thus, in expectation, $\Ellb_{\text{new}}(Z,s)$ is negatively correlated with $\Lb(x,s(Z))$. 
N-DUDE treats $\mathbf{L}_{\text{new}}^\top\mathds{1}_{Z_i}\in\mathbb{R}^{|\mcS|}$ as the target ``pseudo-label'' vector for the mapping to apply at location $i$ and minimizes the objective function,
% in Eq.(7) of \cite{MooMinLeeYoo16}, 
\begin{align}
\mathcal{L}(\mathbf{w}, Z^n,\Pib)\triangleq&\frac{1}{n}\sum_{i=1}^n\mathcal{C}\Big(\mathbf{L}_{\text{new}}^\top\mathds{1}_{Z_i}, \mathbf{p}^k(\mathbf{w},\mathbf{C}_i)\Big).\label{eq:objective}
\end{align}
In (\ref{eq:objective}),  $\mathcal{C}(\mathbf{g},\mathbf{p})
% =-\sum_{i=1}^{|\mcS|} g_i\log p_i
$ for $\mathbf{g}\in\mathbb{R}_{+}^{|\mcS|}$ and $\mathbf{p}\in\Delta^{|\mcS|}$ is the (unnormalized) cross-entropy, and $\mathds{1}_{Z_i}$ is the unit vector for the $Z_i$-th coordinate in $\mathbb{R}^{|\mcZ|}$. 
% Hence, for each data index $i$, $\mathbf{L}_{\text{new}}^\top\mathds{1}_{Z_i}\in\mathbb{R}_+^{|\mcS|}$, which is a \emph{random} vector, is treated as the target ``pseudo-label'' vector for the input (context) $\Cb_i$. 
% From the unbiasedness of $\Ellb$, we can see that $s$ with large 
Note $\mathbf{L}_{\text{new}}^\top\mathds{1}_{Z_i}$ is not necessarily a one-hot vector, and from (\ref{eq:est_loss}) and (\ref{eq:L_max}), the mapping $s$ with larger pseudo-label value should have smaller ``true'' loss \emph{in expectation}.

% as in the case of the usual supervised multi-class classification. 
% that uses $\mathbf{L}_{\text{new}}^\top\mathds{1}_{Z_i}\in\mathbb{R}_+^{|\mcS|}$ as a ``pseudo-label'' for the single-symbol denoiser at the $i$-th location. 
% For learning the parameter $\wb$,
% following as the objective function to minimize in order to learn $\mathbf{w}$:
% in which $\mathcal{C}(\mathbf{g},\mathbf{p})=-\sum_{i=1}^{|\mcS|} g_i\log p_i$ for $\mathbf{g}\in\mathbb{R}_{+}^{|\mcS|}$ and $\mathbf{p}\in\Delta^{|\mcS|}$ stands for the (unnormalized) cross-entropy function, and $\mathds{1}_{Z_i}$ stands for the unit vector for the $Z_i$-th coordinate in $\mathbb{R}^{|\mcZ|}$. Hence, for each data index $i$, $\mathbf{L}_{\text{new}}^\top\mathds{1}_{Z_i}\in\mathbb{R}_+^{|\mcS|}$ is treated as the target label vector for the input (context) $\cb_i$, but the vector is not a unit vector as in the case of the usual supervised multi-class classification.
% Given the objective function in (\ref{eq:objective}) is set, 
% the ordinary back-propagation and variants of mini-batch SGD are used to minimize the objective function. 
% In (\ref{eq:objective}), we highlighited the dependency on $\Pib$ that is required for computing the pseudo-labels. 

% Since the pseudo-label $\Ellb_{\text{new}}(Z,s)$ is negatively correlated with $\Lb(x,s(Z))$ in expectation by design (following from (\ref{eq:L_max}) and the unbiased property of $\Ellb$), the network will tend to assign higher probability for the mapping $s$ that has high $\Ellb_{\text{new}}(Z_i,s)$ value for each $\Cb_i$. 

Once (\ref{eq:objective}) is minimized after sufficient number of iterations, the converged parameter is denoted as $\wb^\star$. 
% Then, for each context $\Cb\in\Cb[k]$, the N-DUDE outputs a probability distribution on the single-letter mappings and chooses 
Then, the single-letter mapping defined by N-DUDE for the context $\Cb\in\Cb[k]$ is expressed as 
% \begin{align}
$s_{k,\texttt{N-DUDE}}(\Cb,\cdot)=\arg\max_{s\in\mcS}\mathbf{p}^k(\wb^\star,\Cb)_s,$
% \end{align}
% \begin{align}
% s_{k,\text{N-DUDE}}(\Cb,\cdot)=\arg\max_{s\in\mcS}\mathbf{p}^k(\wb^\star,\Cb)_s,\label{eq:n_dude dfn}
% \end{align}
% Note $\mathbf{p}^k(\wb^*,\cb)_s$ in (\ref{eq:n_dude dfn}) stands for the $s$-th element of the probability vector $\mathbf{p}^k(\wb^*,\cdot)\in\Delta^{|\mcS|}$. 
and the reconstruction at location $i$ becomes 
\be
\hat{X}_{i,\text{N-DUDE}}(\Cb_i,Z_i)=s_{k,\text{N-DUDE}}(\Cb_i,Z_i).\label{eq:n_dude_recons}
\ee
In summary, Neural DUDE denoises the noisy data after adaptively training the network parameters with the \emph{same} noisy data. \cite{MooMinLeeYoo16} shows N-DUDE significantly outperforms DUDE, has more robustness with respect to $k$, and gets very close to the \emph{optimum} denoising performance for stationary sources. 
% \cite{MooMinLeeYoo16} shows encouraing empirical results including the performance of N-DUDE being robust with respect to $k$.
% , and in this paper, we provide theoretical justification of Neural DUDE.

\noindent\textbf{CUDE \citep{RyuKim18}} borrows the idea of using neural network from N-DUDE, but uses it differently to replace  (\ref{eq:emp}) in the DUDE rule with a neural network learned empirical distribution of $z$ given $\Cb$. Namely, it defines a network $\mathbf{p}_{\text{emp}}(\wb,\cdot):\mcZ^{2k}\rightarrow\Delta^{|\mcZ|}$ and train it by minimizing
$
\frac{1}{n}\sum_{i=1}^n\mathcal{C}(\mathds{1}_{Z_i}, \mathbf{p}_{\text{emp}}(\wb,\Cb_i)).
$
Once the learned parameter $\wb^*$ is obtained, CUDE plugs in $\mathbf{p}_{\text{emp}}(\wb^*,\Cb_i))$ in place of $\hat{\mathbf{p}}_{\text{emp}}(\cdot|\Cb_i)$ in  (\ref{eq:dude_rule}). \citep{RyuKim18} shows  CUDE  outperforms N-DUDE primarily due to the reduced output size, $|\mcZ|$ vs. $|\mcS|$.

% \begin{figure*}[t]
%     \centering
%     \includegraphics[width=1\textwidth]{figures/Architecture.pdf}
%     \caption{N-DUDE, CUDE, ICE}\label{fig:architectue}
% \end{figure*}
% \vspace{-.1in}
% \subsection{Baum-Welch (BW) algorithm for HMM}

\noindent\textbf{Baum-Welch (BW) algorithm \cite{BaumPetrie70} for HMM}
\ \ From the equations (\ref{eq:dude_rule}) and (\ref{eq:objective}),
% (\ref{eq:est_loss})-(\ref{eq:objective}), 
we confirm that all of above three schemes require the exact knowledge on the channel $\Pib$. As mentioned in Introduction, however, the Baum-Welch (BW) algorithm combined with forward-backward (FB) recursion for HMM is a powerful method that can denoise $Z^n$ without requiring such knowledge on the channel. 
% there exist algorithms that can denoise the noisy data $Z^n$ without requiring such knowledge on the channel. The Baum-Welch (BW) algorithm \cite{BaumPetrie70} combined with forward-backward (FB) recursion for HMM is one such powerful method widely used in practice \cite{KroBroMiaSjoetal94,HuaAriJac90}. 
% Namely, by assuming $x^n$ was generated from a Markov chain, BW can estimate the channel transition matrix (\ie, the emission probability) and the state transition probability with a well-known Expectation-Maximization (EM) algorithm. Since the forward-backward recursion (also known as the Viterbi algorithm) provides with the efficient computation of the posterior probability, $p(x_i|Z^n)$, via dynamic programming, the overall process of BW and denoising via maximum \emph{a posteriori} (MAP) estimation is computationally efficient. 
Despite the strength and being widely used in practice \cite{KroBroMiaSjoetal94,HuaAriJac90}, we note the BW based on HMM has some drawbacks, too. 
% While the BW combined with HMM modeling has the strength of computational efficiency and thus is used widely in practice [SR, Bio], it has several drawbacks. 
Firstly, the Markov assumption on the clean $x^n$ may not be accurate; \ie, $x^n$ may not have generated from a Markov source or the assumed order could have a mismatch from the true model. In such cases, the resulting BW and FB recursion based denoising will have poor performance. Secondly, the BW-based channel estimation may suffer from instability with respect to the initialization of the algorithm. In the later sections, we convincingly show that our proposed ICE can reliably estimate the channel, and the ICE combined N-DUDE and CUDE overcome the drawbacks of BW and achieve significant better denoising performance in realistic discrete data.

\section{Iterative Channel Estimation for N-DUDE}
Our ICE algorithm is based on N-DUDE, and 
% it alternates between the following two steps to estimate $\Pib$ and learn $\wb$ jointly until the objective (\ref{eq:objective}) converges. 
Algorithm \ref{alg1} summarizes its pseudo-code. 

\begin{algorithm}[h]
\caption{ICE algorithm for N-DUDE}
\begin{algorithmic}\label{alg1}
    \REQUIRE Noisy data $Z^n$, A N-DUDE model $\mathbf{p}^k(\wb,\cdot):\mathcal{Z}^{2k}\rightarrow\Delta^{|\mcS|}$
    \ENSURE Channel estimate $\hat{\Pib}$, Network parameters $\hat{\wb}$
    \STATE Randomly initialize $\Pib^{(0)}$ and $\wb^{(0)}$ and fix window size $k$. Set $t\leftarrow0$ 
    % \STATE 
    \WHILE{$|\mathcal{L}(\wb,Z^n;\Pib^{(t)})-\mathcal{L}(\wb,Z^n;\Pib^{(t-1)})|>10^{-3}$}
    % \texttt{adf}
        \STATE Compute $\Ellb_{\text{new}}$ in (\ref{eq:L_max}) using $\Pib^{(t)}$.
        % \STATE \texttt{/*Approximate E-step (update $\wb$)*/}
        \STATE \textbf{Approximate E-step (update $\wb$):}\  Obtain $\wb^{(t+1)}$ and the induced posterior for each location $i$, $q(x_i|Z_{i-k}^{i+k};\wb^{(t+1)})$, as following:\vspace{-.05in}
\begin{eqnarray}
\wb^{(t+1)}&=&\underset{\wb}{\arg\min}\ \mathcal{L}(\wb,Z^n;\Pib^{(t)})\label{eq:w_update}\\
q(x_i|Z_{i-k}^{i+k};\wb^{(t+1)})&\triangleq&\sum_{s:s(Z_i)=x_i}\mathbf{p}^k(\wb^{(t+1)},\Cb_i)_s\label{eq:ind_post}
% =&\underset{\wb}{\arg\min}\frac{1}{n}\sum_{i=1}^nD\Big(p_{\Ellb_{\text{new}}}(\cdot|Z_i;\Pib^{(t)}) \ \big\| \ p^k(\cdot|\Cb_i;\wb)\Big)\label{eq:w_update}.
\end{eqnarray}
\vspace{-.05in}
% \STATE We always do a warm-start from the weight of the previous iteration, $\wb^{(t)}$, except for the first iteration, when carrying out the minimization (\ref{eq:w_update}).
        % \STATE Obtain $\wb^{(t+1)}$ and $q(x_i|Z_{i-k}^{i+k};\wb^{(t+1)})$ as in (\ref{eq:w_update}) and (\ref{eq:ind_post}), respectively.
%         \begin{align}
% \wb^{(t+1)}=&\underset{\wb}{\arg\min}\ \mathcal{L}(\wb,Z^n;\Pib^{(t)})\label{eq:w_update}.
% % =&\underset{\wb}{\arg\min}\frac{1}{n}\sum_{i=1}^nD\Big(p_{\Ellb_{\text{new}}}(\cdot|Z_i;\Pib^{(t)}) \ \big\| \ p^k(\cdot|\Cb_i;\wb)\Big)\label{eq:w_update}.
% \end{align}
% \STATE \texttt{/*M-step (update $\Pib$)*/}
        \STATE \textbf{M-step (update $\Pib$):}\ Obtain $\Pib^{(t+1)}$ as 
\begin{align}
\Pib^{(t+1)}(j,k)=\frac{\sum_{i=1}^n\mathds{1}_{\{Z_i=k\}}q(x_i=j|Z_{i-k}^{i+k};\wb^{(t+1)})}{\sum_{i=1}^{n}q(x_i=j|Z_{i-k}^{i+k};\wb^{(t+1)})}.\label{eq:m_step}
\end{align}
\vspace{-.05in}
        \STATE $t\leftarrow t+1$
        
        % \STATE Obtain sequence of mappings $\hat{s}(\cb_i,\cdot)=\mathbf{p}(\mathbf{w},\cb_i)$
    \ENDWHILE
\STATE $\hat{\Pib}=\Pib^{(t)}$.  Do a final update (\ref{eq:w_update}) with $\hat{\Pib},\wb^{(t)}$ and obtain $\hat{\wb}$
% \STATE
\end{algorithmic}
\end{algorithm}

\emph{Remark:} When carrying out the minimization in (\ref{eq:w_update}), we always do a warm-start from the weight of the previous iteration, $\wb^{(t)}$, except for the first iteration. Moreover, (\ref{eq:m_step}) looks very similar to the M-step of BW, and the intuition behind these update formulae is given below.

% contrast, two unique characteristics of N-DUDE enables the iterative estimation. 

% the channel information is used to compute the pseudo-labels (\ref{eq:L_max}) for training the network, hence, an updated channel would lead to an updated network

% denote the estimate of the channel as $\hat{\Pib}$ and the learned network parameters as $\hat{\wb}$. We then 
% 
\subsection{Intuition behind ICE}

% For $Q(x^n;\wb)$ obtained the by the approximate E-step described above, we carry out the M-step as in Baum-Welch (BW). That is, we obtain 
% \be
% \Pib^{(t+1)}=\underset{\Pib}{\arg\max} \sum_{x^n}Q(x^n;\wb^{(t+1)})\log\prod_{i=1}^n\Pib(x_i,Z_i)\nonumber
% \ee 
% which becomes
% \be
% \Pib_{jk}^{(t+1)}=\frac{\sum_{i=1}^n\mathds{1}\{Z_i=k\}p(x_i=j|\Cb_i,Z_i;\wb^{(t+1)})}{\sum_{i=1}^np(x_i=j|\Cb_i,Z_i;\wb^{(t+1)})}\label{eq:m_step}
% \ee
% A detailed derivation of (\ref{eq:m_step}) is given in the Supplementary Material. 

The intuition behind the ICE algorithm lies in the argument of maximum likelihood estimation with variational approximation. That is, we maintain the \emph{stochastic setting} of \cite{Dude} and denote $p_{\mcX}(x^n)$ as the (unknown) prior distribution of the clean sequence $x^n$. Then, we also denote $p(x^n,Z^n;\Pib^{(t)})$ as the joint distribution of $(x^n,Z^n)$ induced from $p_{\mcX}(x^n)$ and $\Pib^{(t)}$, \ie, 
$
p(x^n,Z^n;\Pib^{(t)})=p_{\mcX}(x^n)p(Z^n|x^n;\Pib^{(t)})\label{eq:pi_dependency}=p_{\mcX}(x^n)\prod_{i=1}^n\Pib^{(t)}(x_i,Z_i).\nonumber
$
% Note in (\ref{eq:pi_dependency}), we highlighted the dependency on $\Pib$ since we treat it as unknonw parameters. 
Now, the standard variational lower bound on $\log p(Z^n;\Pib^{(t)})$ becomes
% the log-likelihood of $Z^n$ with $\Pib^{(t)}$ becomes
% Then, we let $p_{X,n}(x^n)\triangleq \prod_{i=1}^np_{X,n}(x_i)$, \ie, assume the empirical joint distribution of $x^n$ as the product of the marginal distributions.  
\begin{align}
% &\log p(Z^n;\Pib^{(t)})
% = \log \sum_{x^n} p(x^n,Z^n;\Pib^{(t)})\nonumber\\
   				 %=&\log \sum_{x^n} Q(x^n) \frac{p(x^n,Z^n;\Pib^{(t)})}{Q(x^n)}\nonumber\\
   				 %\geq 
   				 \sum_{x^n} Q(x^n)\log\frac{p(x^n,Z^n;\Pib^{(t)})}{Q(x^n)}=\log p(Z^n;\Pib^{(t)})-D_{KL}\Big(Q(x^n)\|p(x^n|Z^n;\Pib^{(t)})\Big)\label{eq:lower bound}
\end{align}
in which $Q(x^n)$  stands for any probability distribution on $x^n$, and $D_{KL}(\cdot\|\cdot)$ is the Kullback-Leibler divergence. 
% , and the inequality follows from the Jensen's inequality. 
For a fixed $\Pib^{(t)}$, one can easily see that $Q(x^n)$ that maximizes (\ref{eq:lower bound}) becomes $Q(x^n)=p(x^n|Z^n;\Pib^{(t)})$,
% \be
% Q(x^n)=p(x^n|Z^n;\Pib^{(t)}),\label{eq:posterior}
% \ee
the posterior of $x^n$ given $Z^n$ derived from the joint distribution $p(x^n,Z^n;\Pib^{(t)})$. Note this is equivalent to the E-step in EM algorithm and BW for HMM, in particular. Given above
$Q(x^n)=p(x^n|Z^n;\Pib^{(t)})$, 
% (\ref{eq:posterior}), 
the standard M-step in BW solves 
\begin{align}
    \underset{\mathbf{\Pi}}{\arg\max}\sum_{x^n}Q(x^n)\log \frac{p(x^n,Z^n;\Pib)}{Q(x^n)}\label{eq:em_m_step}
\end{align}
to obtain $\Pib^{(t+1)}$, which results in using the marginal posterior $p(x_i|Z^n;\Pib^{(t)})$ in the update formula as in (\ref{eq:m_step}). Note in HMM, computing such marginal posterior can be done efficiently thanks to the Markov assumption on $x^n$.
% In HMM, obtaining the marginal of the posterior, $p(x_i|Z^n;\Pib^{(t)})$, can be done efficiently thanks to the Markov assumption on $x^n$.
% via the Markov assumption on $x^n$, the marginal posterior $p(x_i|Z^n;\Pib)$ can be efficiently obtained via forward-backward recursion. 
However, in our universal setting, in which no distributional assumption is made on $x^n$, obtaining the posterior $p(x^n|Z^n;\Pib^{(t)})$ or its marginal becomes intractable. 

% \noindent \textbf{Approximate E-step (update $\wb$):} 
Therefore, instead of $Q(x^n)=p(x^n|Z^n;\Pib^{(t)})$, our ICE uses 
\begin{align}
    Q(x^n)=\tilde{Q}(x^n;\Pib^{(t)})\triangleq\prod_{i=1}^nq(x_i|Z_{i-k}^{i+k};\wb^{(t+1)})\label{eq:q_approx_e_step}
\end{align}
as the \emph{approximate E-step} expecting 
$D_{KL}(\tilde{Q}(x^n;\Pib^{(t)})\|p(x^n|Z^n;\Pib^{(t)}))$
becomes small such that the lower bound (\ref{eq:lower bound}) becomes sufficiently tight. The reasoning for this approximation follows from the strong empirical performance of N-DUDE that suggests the induced posterior (\ref{eq:ind_post})
becomes a good approximation for $p(x_i|Z^n;\Pib^{(t)}).$
Namely, note that with the true channel $\Pib$, the optimum reconstruction for the $i$-th location for minimizing the average loss becomes
\be
\hat{X}_{i,\text{opt}}(Z^n)=\arg\min_{\hat{x}\in\hat{\mathcal{X}}}\sum_{x=1}^{|\mathcal{X}|}\Lb(x,\hat{x})p(x_i=x|Z^n;\Pib),\label{eq:opt_recon}
\ee
which depends on the marginal posterior, $p(x_i|Z^n;\Pib)$. Now, as shown in \citep[Figure 2]{MooMinLeeYoo16}, the reconstruction of N-DUDE, (\ref{eq:n_dude_recons}),
% \be
% \hat{X}_{i,\text{N-DUDE}}(Z^n)=s_{k,\text{N-DUDE}}(\Cb_i,Z_i),
% \ee
which applies the mapping obtained by $\mathbf{p}^k(\wb^\star,\Cb)$,
% (\ref{eq:n_dude dfn})
% $$
% s_{k,\text{N-DUDE}}(\Cb,\cdot)=\arg\max_{s\in\mcS}\mathbf{p}^k(\wb^\star,\Cb)_s,
% $$
can attain the optimum performance of (\ref{eq:opt_recon}) for sufficiently large $k$. Therefore, we can expect the induced posterior
$
q(x_i|Z_{i-k}^{i+k};\wb^\star)=\sum_{s:s(Z_i)=x_i}\mathbf{p}^k(\wb^\star,\Cb)_s
$
approximates $p(x_i|Z^n;\Pib)$ sufficiently well for large $k$. From above reasoning, our M-step, (\ref{eq:m_step}), follows from solving (\ref{eq:em_m_step}) with $Q(x^n)$ given by the approximate E-step in (\ref{eq:q_approx_e_step}), and the precise derivation is given in the Supplementary. 
% In our experiments, we give a concrete validation for this intuition.

\emph{Remark:} One important point to make is that it is \emph{not} possible to devise an iterative channel estimation scheme like ICE based on CUDE  \cite{RyuKim18}. The reason is because in CUDE, the channel $\Pib$ only occurs in the final denoising rule (\ref{eq:dude_rule}), and learning the neural network has nothing to do with $\Pib$. In N-DUDE, however, $\Pib$ is used to compute the pseudo-labels (\ref{eq:L_max}) for training the network, hence, an updated channel would lead to an updated network. Moreover, from the updated network, N-DUDE can naturally induce the marginal posterior (\ref{eq:ind_post}) to carry out the M-step to update $\Pib$.

\section{Experimental Results}\label{sec:experiment}

% In this section, we carry out extensive experiments using synthetic data, real binary images, and Oxford nanopore MinION DNA sequence data to show the effectiveness and robustness of our \ice algorithm. 
% % show the results of \icendude for synthetic data which is second-order Hidden Markov Process, real binary images, and real Oxford nanopore MinoION DNA sequence data. 
% All the experiments were done with Python 3.6 and Keras 
% % package
% % (\texttt{http://keras.io}) 
% with Tensorflow backend. 

\noindent\textbf{Data and training details} \ We carry out extensive experiments using synthetic data, real binary images, and Oxford nanopore MinION DNA sequence data to show the effectiveness and robustness of our ICE algorithm. 
% show the results of \icendude for synthetic data which is second-order Hidden Markov Process, real binary images, and real Oxford nanopore MinoION DNA sequence data. 
All the experiments were done with Python 3.6 and Keras 
% package
% (\texttt{http://keras.io}) 
with Tensorflow backend. 
For the \emph{approximate E-step} in (\ref{eq:w_update}), we used the Adam optimizer \cite{KinBa15} with default setting to minimize the objective function. The number of epochs for each iteration was set to 10 for synthetic and real binary image data and 20 for DNA data. The initial learning rate for the first iteration was $10^{-3}$, then from the second iteration, we used $10^{-4}$. As shown below in Figure \ref{fig:loss_PI_bar}, the objective function quickly converges after a few iterations, hence, we stopped the estimation process after the third iteration in all of our experiments. For the network architectures, we used 3 fully connected layers with 40  nodes and 160  nodes for synthetic/binary image data and DNA data, respectively. 

\subsection{Synthetic data}\label{eq:synthetic}
\begin{figure*}[ht]
    \centering
    \includegraphics[width=1\textwidth]{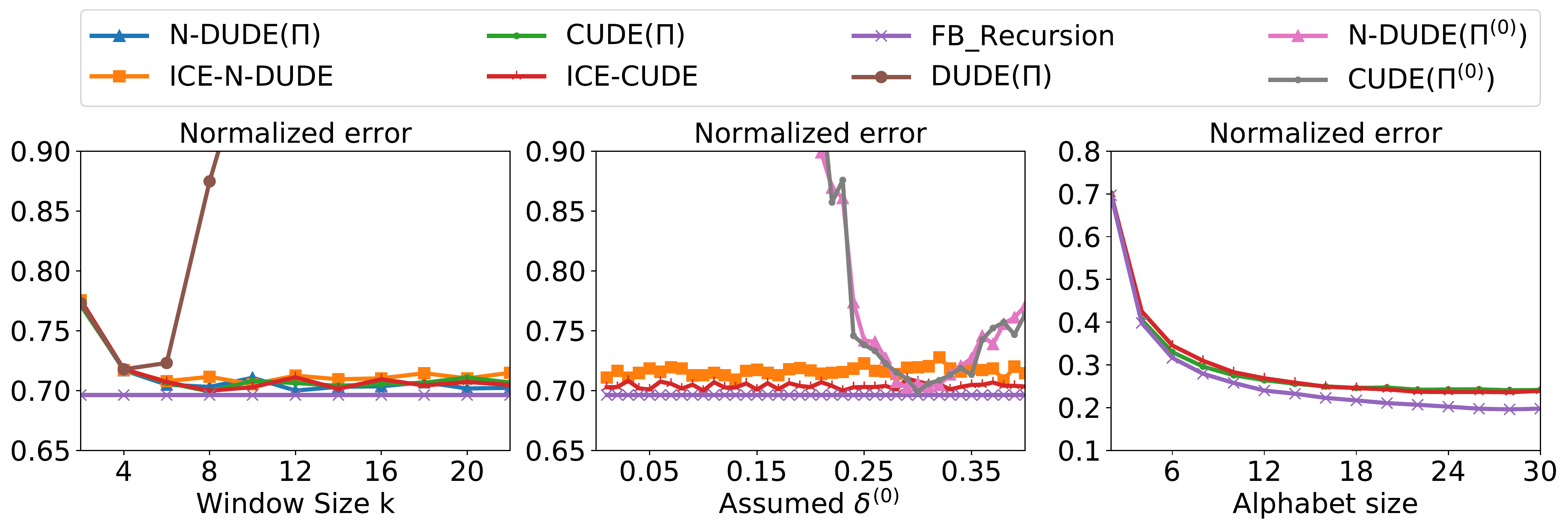}
    
    \caption{\small{Denosing results for HMM with respect to window size $k$ (left), assumed $\delta^{(0)}$ (center), and the alphabet size $|\mcX|$ (right). The vertical axes for three figures correspond to the normalized error rates.}}\label{fig:HMM_error_k_delta_alpha}
\end{figure*}

First, we carry out the experiment on synthetic data to validate the performance of ICE. Following \cite{MooMinLeeYoo16}, we generated the clean binary data from a binary symmetric Markov chain (BSMC) with transition probability $\alpha=0.1$. The data was corrupted by a binary symmetric channel (BSC) $\Pib$ with cross-over probability $\delta=0.3$ to result in the noisy sequence $Z^n$, which becomes a hidden Markov process. The length of the sequence was set to $n=10^6$, and the Hamming loss was used to set the Bit Error Rate (BER) as $\frac{1}{n}\sum_{i=1}^{n}\mathds{1}\{x_{i}\neq\hat{x}_{i}\}$.
% Then, we divided this by $\delta$. 
We report the normalized error, obtained by dividing BER with $\delta$. 
% obtained by 
The denoising results on this data are given in the left and center plots of Figure \ref{fig:HMM_error_k_delta_alpha}, in which the left shows with respect to the window size $k$ and the center shows with respect to the initially assumed $\Pib^{(0)}$. Note since $Z^n$ is a hidden Markov process in this case, the FB-Recursion that \emph{knows} $\Pib$ and the state transition probability $\alpha$ can achieve the optimum denoising performance, shown as purple lines (\ie, lower bounds) in the figures. 

In Figure \ref{fig:HMM_error_k_delta_alpha} (left), DUDE($\Pib$), N-DUDE($\Pib$) and CUDE($\Pib$) stand for the results of the three schemes that exactly know the true channel $\Pib$. We can confirm the universality of those methods since they almost achieve the optimum performance, while not knowing the source is a Markov. Also, N-DUDE($\Pib$) and CUDE($\Pib$) are much more robust with respect to $k$ than DUDE($\Pib$). For our ICE, we initialized $\Pib^{(0)}$ as BSC with crossover probablility $\delta^{(0)}=0.1$, and we observe ICE-N-DUDE and ICE-CUDE, which \emph{plug-in} the estimated channel by ICE to N-DUDE and CUDE, respectively, work very well and essentially achieve the same performances as their counterparts that know $\Pib$. We stress that this
is a nontrivial result since ICE just observes $Z^n$ and provides an accurate enough estimation of $\Pib$ to achieve the optimum performance, only with the independent noise assumption.

In Figure \ref{fig:HMM_error_k_delta_alpha} (center), we show the robustness of ICE with respect to varying initially assumed $\delta^{(0)}$, while fixing the window size $k=16$. In the figure, N-DUDE($\Pib^{(0)}$) and CUDE($\Pib^{(0)}$) are the schemes that run with $\Pib^{(0)}$, which can be potentially mismatched with the true $\Pib$, and we clearly see they become very sensitive to the mismatch of the assumed $\Pib^{(0)}$. In contrast, ICE becomes extremely robust to the initial $\Pib^{(0)}$ such that both ICE-N-DUDE and ICE-CUDE almost achieve the optimum performance regardless of the initially assumed $\delta^{(0)}$. 

% We also see that ICE-CUDE is slightly be

% This robustness of ICE is again confirmed in the later experiments. 

% assume channel matrix as initial channel $\Pib^{(0)}$ which is quite different from true channel $\Pib$. In this experiment, we ran N-DUDE($\Pib^{(0)}$), CUDE($\Pib^{(0)}$), ICE-N-DUDE and ICE-CUDE with $k=16$ multiple times with different initial $\delta$ values; note that N-DUDE($\Pib^{(0)}$) and CUDE($\Pib^{(0)}$) clearly becomes sensitive to the mismatched $\delta$, but ICE-N-DUDE and ICE-CUDE becomes extremely robust showing the effectiveness of ICE. This robustness of ICE is also confirmed the later experiments. 

In Figure \ref{fig:HMM_error_k_delta_alpha} (right), we also investigate the performance of ICE with respect to the alphabet size of data. We increased the alphabet size of the Markove source, $|\mcX|$ (and $|\mcZ|$), from 2 to 30, and for each case, the transition probabilities from a state to others were set to be uniform as $0.1/(|\mcX|-1)$. $n$ was $5\times 10^6$, and the true $\Pib$ was set such that $\Pib(i,i)=0.7$ and $\Pib(i,j)=0.3/(|\mcZ|-1)$ for $i\neq j$. The initial $\Pib^{(0)}$ for ICE was set such that $\Pib^{(0)}(i,i)=0.9$ and $\Pib^{(0)}(i,j)=0.1/(|\mcZ|-1)$ for $i\neq j$. We compared CUDE($\Pib$), ICE-CUDE and FB-Recursion, and again, ICE-CUDE performs almost as well as CUDE($\Pib$), robustly over the alphabet sizes. The gap from FB-Recursion is primarily due to the fixed $n$, and we beleive it will close once $n$ grows with the alphabet size.

\subsection{Binary image}

Now, we move on to the experiments using more realistic binary images as clean data. We tested on two datasets: PASCAL and Standard. PASCAL consists of 50 binarized grayscale images that we obtained from PASCAL VOC 2012 dataset \cite{pascalvoc12}, and Standard consists of 8 binarized standard images that are widely used in image processing, \{\texttt{Barbara}, \texttt{Boat}, \texttt{C.man}, \texttt{Couple}, \texttt{Einstein}, \texttt{fruit}, \texttt{Lena}, \texttt{Peppers}\}. We tested with three noise levels and applied non-symmetric channels with average noise levels of $0.1$, $0.2$, and $0.3$, and the exact $\Pib$'s are given in the Supplementary Material. As in \cite{MooMinLeeYoo16}, we raster scanned the images and converted them to 1-D sequences.

% Each binary image was again corrupted by a non-symmetric $\Pib$ with several noise levels, 10\%, 20\% and 30\%.

In Table \ref{table:binary_error}, we compare the normalized errors of ICE-N-DUDE and ICE-CUDE with BW that assume the images are Markov. BW\_1st, BW\_2nd, and BW\_3rd correspond to BW with various Markov order assumptions. N-DUDE($\Pib$) and CUDE($\Pib$), which are known to achieve the state-of-the-art for binary image denoising  with \emph{known} channel, are shown as lower bounds. We fixed the window size to $k=50$ for all neural network based schemes.
% and estimates the unknown channel via EM for state estimation (denoising). 
For both ICE and BW, we set the initial $\Pib^{(0)}$ as BSC with $\delta^{(0)}=0.1$ for all noise levels. For the exact procedure of running ICE and training denoisers with multiple images, we refer to the Supplementary Material.

\begin{table}[th]
\small
\centering
\caption{Denoising results for real binary images.
% The assumed channel is BSC with $\delta=0.1$ for all noise level experiments. 
}\vspace{-0.0in}\label{table:binary_error}
\resizebox{0.8\linewidth}{!}{\begin{tabular}{|c|c|c|c|c|c|c|}
\hline
Noise level                     & \multicolumn{2}{c|}{$\delta=0.1$} & \multicolumn{2}{c|}{$\delta=0.2$} & \multicolumn{2}{c|}{$\delta=0.3$} \\ \hline
Methods \textbackslash{}Dataset & PASCAL          & Standard        & PASCAL          & Standard        & PASCAL          & Standard        \\ \hline \hline
BW\_1st                        & 0.4294          & 0.5469          & 0.3508          & 0.4726          & 0.5088          & 0.5716          \\ \hline
BW\_2nd                        & 0.4770          & 0.6342          & 0.4025          & 0.5149          & 0.3829          & 0.5020          \\ \hline
BW\_3rd                        & 0.5996          & 1.0943          & 0.5619          & 0.7523          & 0.5277          & 0.7415          \\ \hline
ICE-N-DUDE                     & 0.3516          & 0.4120          & 0.3223          & 0.3771          & \textbf{0.3429} & 0.4286          \\ \hline
ICE-CUDE                       & \textbf{0.3512} & \textbf{0.4038} & \textbf{0.3205} & \textbf{0.3712} & 0.3438          & \textbf{0.4266} \\
\hline \hline
% N-DUDE ($\Pib^{(0)}$)          & 0.3341          & 0.3885          & 0.7793          & 0.9161          & 1.0000          & 0.9999          \\ \hline
% CUDE ($\Pib^{(0)}$)            & 0.3330          & 0.3889          & 0.7410          & 0.8880          & 0.9999          & 0.9999          \\ \hline \hline
N-DUDE ($\Pib$)                & 0.3540          & 0.3981          & 0.3300          & 0.3826          & 0.3494          & 0.4524          \\ \hline
CUDE ($\Pib$)                  & 0.3259          & 0.3748          & 0.3171          & 0.3684          & 0.3396          & 0.4245          \\ \hline
\end{tabular}}
\end{table}
In the table, we see that both ICE-N-DUDE and ICE-CUDE significantly outperform all three BW methods for all noise levels and datasets, and they  get very close to N-DUDE($\Pib$) and CUDE($\Pib$), respectively. We also confirm the superiority of CUDE over N-DUDE, as claimed in \cite{RyuKim18}. 
% The reason for this would be again due to the local robustness of \ndude around the true $\Pib$, potentially yielding better results for slightly perturbed channel. 
Moreover, note the BW schemes become sensitive depending the noise level and dataset; \ie, for $\delta=0.1,0.2$, BW\_1st performs the best, while for $\delta=0.3$, BW\_2nd is superior. Such difficulty of accurately determining the best order of HMM for a given dataset is one of the main drawbacks of BW method. 
% This is one of the main drawbacks of B 
% The main drawback of BW is the difficulty in accurately determining the best order (\ie, order estimation) for a given dataset. 
% that it is difficult to accurately determine the best order (i.e., order estimation) for the given data. 
On the contrary, ICE-N-DUDE and ICE-CUDE work universally well for all sources and noise levels, and neither the clean source modeling nor the true channel $\Pib$ was necessary.

% does not require the modeling on the clean data and robustly achieves good denoising performance for all cases. 

% More important thing is that N-DUDE($\Pib^{(0)}$) and CUDE($\Pib^{(0)}$) perform quite poor when true noise level is 0.2, 0.3. However, ICE-N-DUDE and ICE-CUDE is highly robust to noise level, which shows that ICE resolves the channel uncertainty effectively.

% the \texttt{Baum-Welch}s and it is highly close to \ndude which knows the exact channel. The assumption of markov chain order makes \texttt{Baum-Welch} sensitive. Hence, \texttt{Baum-Welch} has various errors depending on the order.
% On the other hand, since our methods, \ndude and \icendude, are universal denoiser, in this part, our model goes far beyond Baum-Welch. 

\subsection{DNA sequence}

We now apply ICE to DNA sequence denoising and mainly follow the experimental setting of \citep[Section 5.3]{MooMinLeeYoo16};
%\cite[Section 5.3]{MooMinLeeYoo16};
namely, we obtained 16S rDNA reference sequences for 20 species and randomly generated noiseless template reads $x^n$ of length $n=2,469,111$. Then, we used the same $\Pib$ in
\cite{MooMinLeeYoo16}, which had $20.375\%$ average error rate, to corrupt $x^n$ and obtain $Z^n$. The true (asymmetric) $\Pib$ is given in the Supplementary Material. For ICE, the initial $\Pib^{(0)}$ was assumed to be $\Pib^{(0)}(i,i)=0.6$ and $\Pib^{(0)}(i,j)=0.4/(|\mcZ|-1)$ for $i\neq j$. We also abuse the notation and define $\delta^{(0)}\triangleq \sum_{j\neq i}\Pib^{(0)}(i,j)$ which becomes $0.4$ for all $i$. Note as shown in \citep{LeeMooYooWei16,MooMinLeeYoo16}, DUDE($\Pib$) and N-DUDE($\Pib$) can achieve the state-of-the-art for DNA sequence denoising as well. 

\vspace{-.1in}
\begin{figure*}[h]
    \centering
    \subfigure[Error rate with respect to window size k]{\label{fig:DNA_k}

    \includegraphics[width=0.40\textwidth]{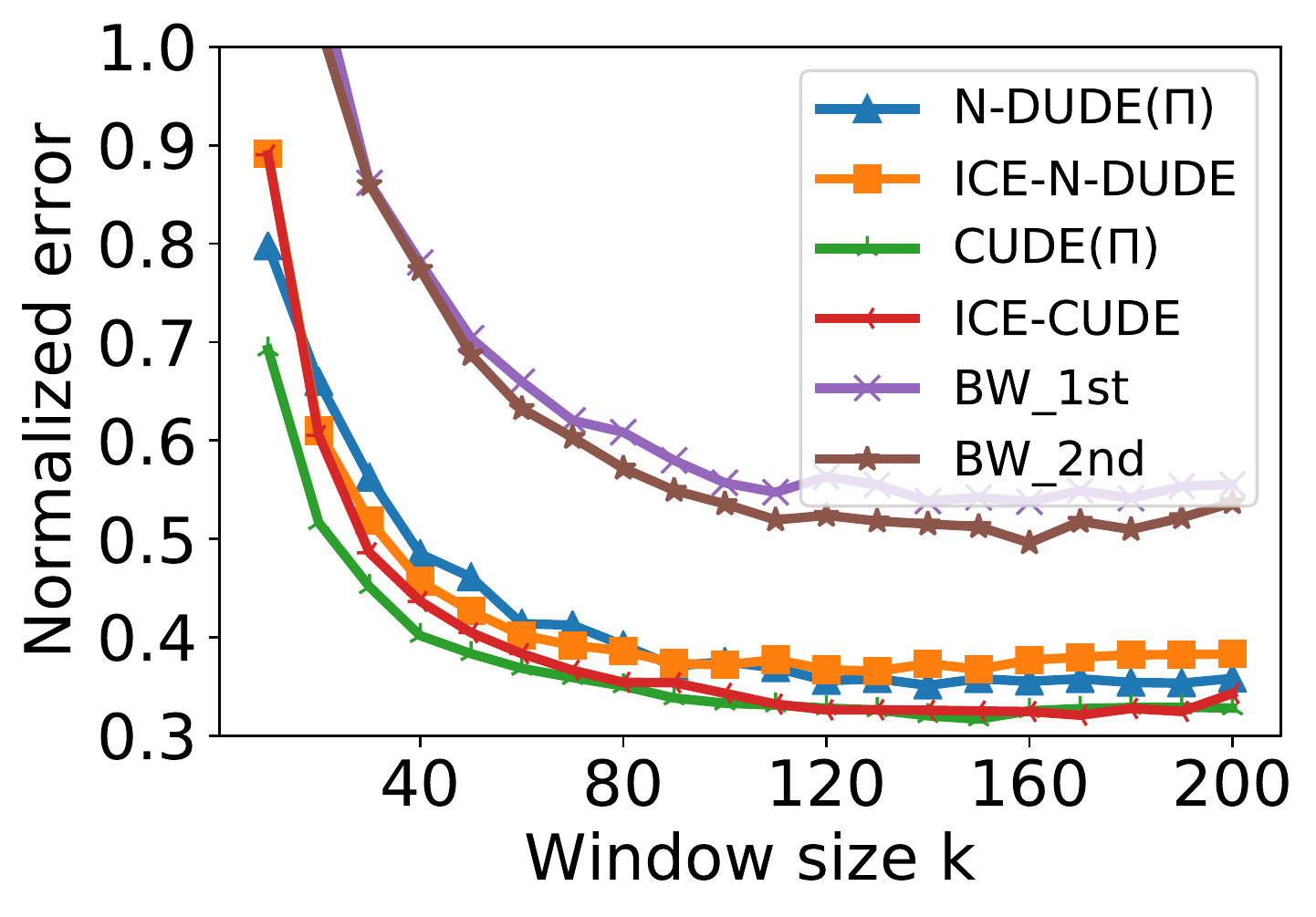}}
    \subfigure[Boxplots of error rates with varying initial $\delta^{(0)}$.]{\label{fig:DNA_delta_box}

    \includegraphics[width=0.51\textwidth]{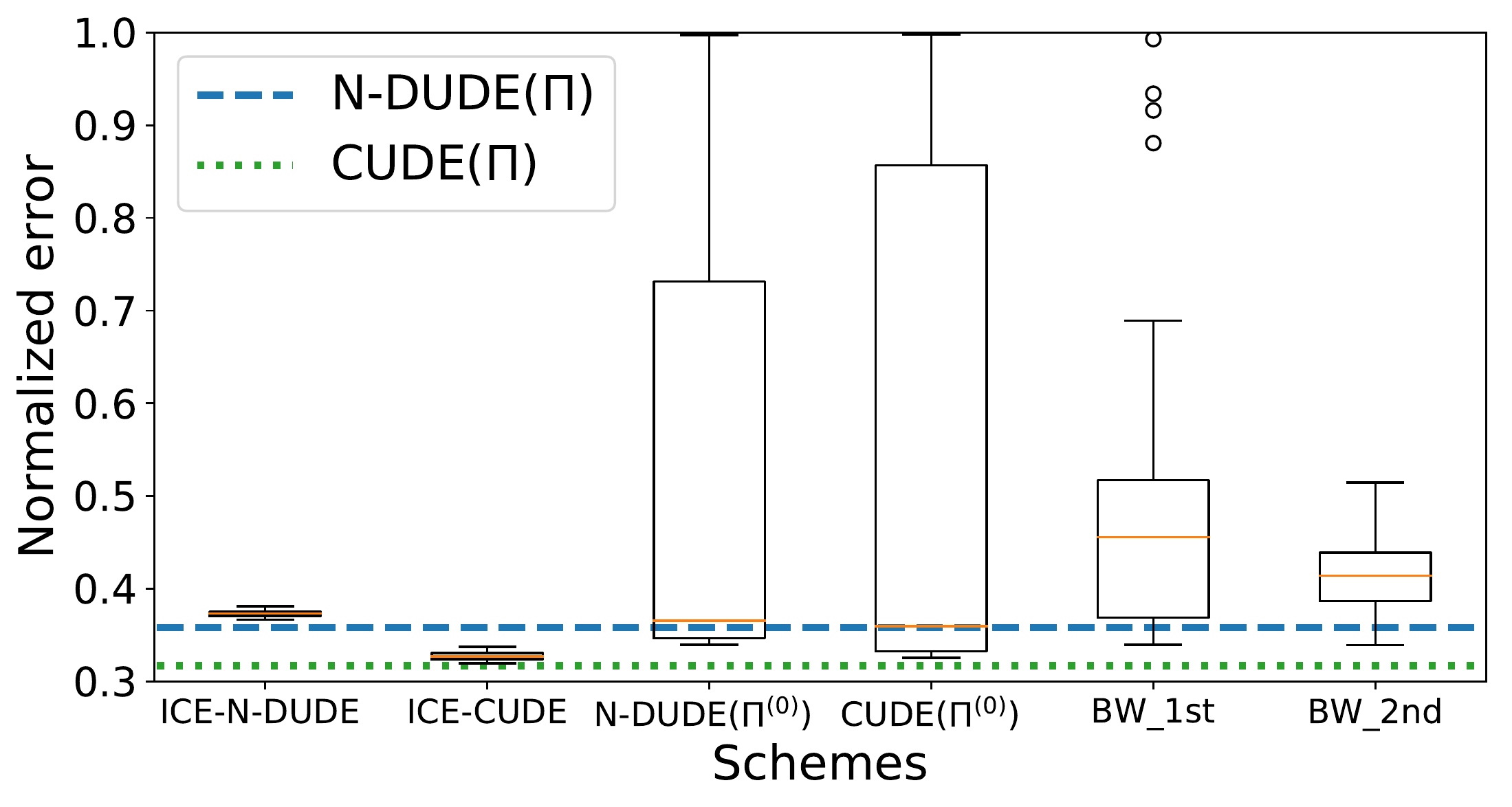}}
    \caption{Denosing results for real DNA data with respect to window size $k$ \& initial $\Pib^{(0)}$.}\label{fig:DNA_error}
\end{figure*}
Figure \ref{fig:DNA_k} shows the denoising results with varying window size $k$, and Figure \ref{fig:DNA_delta_box} shows the boxplots of normalized errors with window size $k=150$ and varying initial $\delta^{(0)}$'s within the range $0.01\sim0.40$ (40 samples).
% For this experiment, the original output dimension of 256 for \ndude and \icendude did not perform well at all, hence we report the results for the reduced dimension cases, 16 mappings and 5 mappings (shown in the parenthesis of the legends in Figure \ref{fig:DNA_k}). 
BW\_1st and BW\_2nd completely failed for this experiment, resulting in the normalized error rate of $1.440$ and $3.153$, respectively. Thus, we could not include them in the figure, and we instead included the results of a hybrid method, i.e, running N-DUDE with BW estimated channels. BW\_1st and BW\_2nd in the figure stand for such hybird of BW with N-DUDE. 
% BW\_3rd and higher orders did not given meaningful results. 
% To show the effect of output dimension reduction described above, we have also added an experiment to compare our models with many mapping models. We labeled a model with 5 mappings as N-DUDE (5) and a model with 16 mappings as N-DUDE (16). As in previous experiments, we assumed that the channel is symmetric. In this case, both \bwone and \bwtwo completely failed, resulting in the normalized error rate of $1.440$ and $3.153$, respectively, thus, we could not include them in the figure. 
% Instead, we decided instead to use the channels from \bwone and \bwtwo with N-DUDE. 
% Therefore, \bwone (5) and \bwtwo (5) represent the N-DUDE (5) model with the channel estimated by \bwone and \bwtwo, respectively. Note comparing with \cite{MooMinLeeYoo16}, \ndude and \icendude can achieve much better error rates compared to \dude of \cite{Dude}, which knows $\Pib$.

Paralleling the results in the previous sections, we observe from Figure \ref{fig:DNA_k} that ICE-N-DUDE and ICE-CUDE get very close to N-DUDE($\Pib$) and CUDE($\Pib$), respectively, and significantly outperform the BW hybrid methods, as $k$ increases. This shows the accuracy and effectiveness of ICE; its channel estimation quality is much better than BW when the underlying $x^n$ is far from being a Markov and while it is based on N-DUDE, the estimated channel can be readily plugged-in to other scheme like CUDE. 
% Particularly, ICE-CUDE performs almost the same as CUDE($\Pib$) when $k$ is larger than 120. 
We believe this is quite a strong result, since ICE-CUDE can remove almost 70\% of noise solely based on $Z^n$ and with no other information on the noise and the clean source. 
% it shows the channel estimated by ICE (which is based on N-DUDE) ca
% Quantitatively, the difference between ICE-N-DUDE and N-DUDE($\Pib$) is only about up to 3.9\% for each $k$, which is quite an impressive result for ICE-N-DUDE. 
% And for CUDE($\Pib$) and ICE-CUDE, they performs almost same when $k$ is above 120, and its performance is quite better than N-DUDE($\Pib$) and ICE-N-DUDE.
% have a performance difference of about 3.9\% for each k with minimum error rate and the largest relative difference with respect to assumed $\delta$ is only 6.90\%. 
Moreover, in Figure \ref{fig:DNA_delta_box}, we see that ICE-N-DUDE and ICE-CUDE are extremely robust with respect to the initial $\delta^{(0)}$, while the mismatched N-DUDE($\Pib^{(0)}$) and CUDE($\Pib^{(0)}$) completely fails for wrong initializations. Moreover the BW hybrid methods also show large variance, mainly due to the sensitivity of the channel estimation quality of BW with respect to the initially assumed $\delta^{(0)}$. 
% We believe this robustness of ICE is an attractive nature in practice. 

% this is another strong point of ICE in addition to its strong mean denoising performance.
% the error rate variance of \icendude (5) w.r.t assumed $\delta$ is fairly small, which means it is highly robust to assumed $\delta$. We believe this is quite a strong result as our method essentially removes the channel uncertainty for denoising complex realistic source, while the standard HMM-based BW algorithm with N-DUDE (5), \bwone (5) and \bwtwo (5), completely fails.
% We also see \ndude(5) and \icendude(5) are much better than the counterparts with 16 mappings, suggesting the output dimension reduction to $|\hat{\mcX}|+1$ is much more favorable for larger alphabets.  

% For the comparison of \ndude (5) and \ndude (16), \ndude (16) always has higher error rate than \ndude (5) and so does \icendude (16). Through this result, we show that the reduction of output dimension has significant performance improvement in both denoising and estimation.

\subsection{Convergence \& estimation analyses}

\begin{figure*}[h]
    \centering
    \includegraphics[width=1\textwidth]{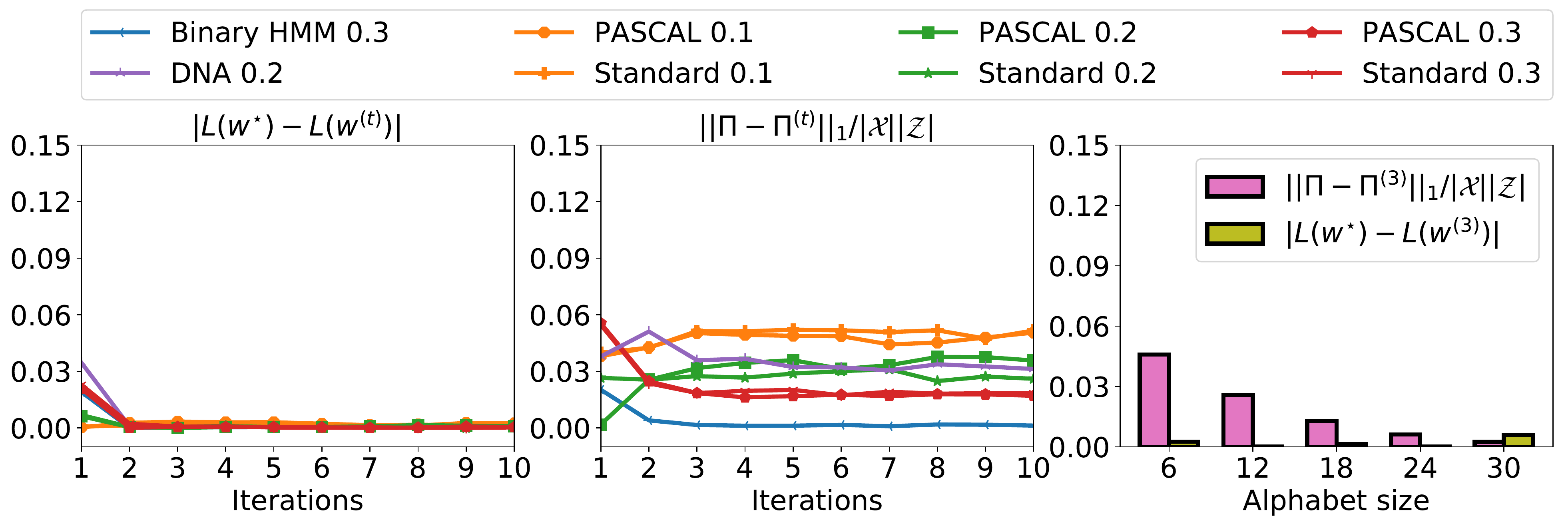}\vspace{-.1in}
    \caption{Convergence and estimation performance analysis}\label{fig:loss_PI_bar}
\end{figure*}

We now give a closer analyses on the channel estimation performance of ICE for all of our experiments given in above sections. 
Figure \ref{fig:loss_PI_bar} shows the following two metrics; (a) $|\mathcal{L}(\wb^\star,Z^n;\Pib)-\mathcal{L}(\wb^{(t)},Z^n;\Pib)|$ and (b) $\|\Pib-\Pib^{(t)}\|_1/|\mcX||\mcZ|$.
% with respect to the iteration of ICE, $t$, and the alphabet sizes.
% \begin{compactitem}
% % \item $\frac{1}{n}\sum_{i=1}^nD_{KL}(q(x_i|Z_{i-k}^{i+k};\Pib^{(t)})||p(x_{i}|Z^{n};\Pib)\big)$
% \item $|\mathcal{L}(\wb^\star,Z^n;\Pib)-\mathcal{L}(\wb^{(t)},Z^n;\Pib)|$
% \item $\|\Pib-\Pib^{(t)}\|_1/|\mcX||\mcZ|$
% \end{compactitem}
The first metric shows the difference between the value of the objective function (\ref{eq:objective}) for N-DUDE($\Pib$) parameter $\wb^\star$ and for the model $\wb^{(t)}$ after each approximate E-step of ICE. Note (\ref{eq:objective}) is computed with the true $\Pib$. The second metric is the normalized $L_1$-norm of $\Pib-\Pib^{(t)}$, which directly measures the channel estimation accuracy. 
% \begin{table}[th]
% \small
% \centering
% \caption{
% % The assumed channel is BSC with $\delta=0.1$ for all noise level experiments. 
% }\vspace{-0.0in}\label{table:binary_error}
% \resizebox{0.8\linewidth}{!}{\begin{tabular}{|c|c|c|c|c|c|c|}
% \hline
% Metrics \textbackslash{}$|\mcX|$                                                             & 6     & 12    & 18    & 24    & 30    \\ \hline
% $|\mathcal{L}(\wb^\star,Z^n;\Pib)-\mathcal{L}(\wb^{(3)},Z^n;\Pib)|$                         & 0.002 & 0.000 & 0.001 & 0.000 & 0.006 \\ \hline
% $\|\Pib-\Pib^{(3)}\|_1/|\mcX||\mcZ|$                                                        & 0.045 & 0.025 & 0.012 & 0.006 & 0.002 \\ \hline
% $\frac{1}{n\times |\mcX|}\sum_{i=1}^nD_{KL}(q(x_i|Z_{i-k}^{i+k};\Pib^{(t)})||p(x_{i}|Z^{n};\Pib)\big)$   & 0.160 & 0.113 & 0.099 & 0.078 & 0.072 \\ \hline 
% \end{tabular}}
% \end{table}

The first two figures show the two metrics for all experiments, respectively, except for the large alphabet Markov source case, with respect to the iteration $t$ of ICE. For metric (a), we observe that the difference becomes very small after just a few iterations for all cases. 
% This means that $\wb^{(t)}$ achieves  essentially the same loss value as that of N-DUDE($\Pib$). 
The result suggests that $\wb^{(t)}$ of ICE and $\wb^\star$ become indistinguishable from the perspective of objective function value, hence, it justifies the good performance of ICE-N-DUDE in denoising experiments. 
% The second metric is the normalized $L_1$ norm of the channel estimation error matrix, $\Pib-\Pib^{(t)}$. 
For metric (b), we observe the estimation errors also become small and stable with respect to the iteration $t$. The excellent denoising performance of ICE-CUDE, which just plugs-in the estimated channel to CUDE, confirms that such level of error is tolerable and has negligible effect in denoising. Moreover, the estimation errors seem to get smaller for larger noise levels. The third figure shows the two metrics at iteration 3 for the large alphabet Markov source case in Section \ref{eq:synthetic}. Again, they both become extremely small confirming the decent estimation quality of ICE for large alphabet case.

\section{Conclusion}
In this paper, we proposed a novel iterative channel estimation method for removing the known channel assumption of N-DUDE. The resulting ICE-N-DUDE and ICE-CUDE achieved excellent denoising performance for various different types of data, without \emph{any} knowledge on the channel and the clean source. For future work, we plan to extend this approach to more general settings, e.g., to general state estimation beyond denoising and to continuous-alphabet case.  
% on the convergence of our method. Furthermore, we believe our method can be extended to a more general universal state estimation problem beyond denoising and has a potential to get widely applied to many applications and excel the popular BW and HMM-based methods. 

\bibliographystyle{abbrvnat}
\bibliography{bibfile}

\begin{thebibliography}{19}
\providecommand{\natexlab}[1]{#1}
\providecommand{\url}[1]{\texttt{#1}}
\expandafter\ifx\csname urlstyle\endcsname\relax
  \providecommand{\doi}[1]{doi: #1}\else
  \providecommand{\doi}{doi: \begingroup \urlstyle{rm}\Url}\fi

\bibitem[Baum et~al.(1970)Baum, Petrie, Soules, and Weiss]{BaumPetrie70}
L.~Baum, T.~Petrie, G.~Soules, and N.~Weiss.
\newblock A maximization technique occuring in the statistical analysis of
  probabilistic functions of {M}arkov chains.
\newblock \emph{Annals of Mathematical Statistics}, 41\penalty0 (164-171),
  1970.

\bibitem[Ephraim and Merhav(2002)]{EphMer02}
Y.~Ephraim and N.~Merhav.
\newblock Hidden markov processes.
\newblock \emph{{IEEE} Trans. Inform. Theory}, 48\penalty0 (6):\penalty0
  1518--1569, 2002.

\bibitem[Everingham et~al.()Everingham, Van~Gool, Williams, Winn, and
  Zisserman]{pascalvoc12}
M.~Everingham, L.~Van~Gool, C.~K.~I. Williams, J.~Winn, and A.~Zisserman.
\newblock The {PASCAL} {V}isual {O}bject {C}lasses {C}hallenge 2012 {(VOC2012)}
  {R}esults.
\newblock
  \texttt{http://www.pascal-network.org/challen\\ges/VOC/voc2012/workshop/index.html}.

\bibitem[Gemelos et~al.(2006{\natexlab{a}})Gemelos, Sigurjonsson, and
  Weissman]{GemSigWes06}
G.~Gemelos, S.~Sigurjonsson, and T.~Weissman.
\newblock Algorithms for discrete denoising under channel uncertainty.
\newblock \emph{{IEEE} Trans. Signal Process.}, 54\penalty0 (6):\penalty0
  2263--2276, 2006{\natexlab{a}}.

\bibitem[Gemelos et~al.(2006{\natexlab{b}})Gemelos, Sigurjonsson, and
  Weissman]{GemSigWes06a}
G.~Gemelos, S.~Sigurjonsson, and T.~Weissman.
\newblock Universal minimax discrete denoising under channel uncertainty.
\newblock \emph{{IEEE} Trans. Inform. Theory}, 52:\penalty0 3476--3497,
  2006{\natexlab{b}}.

\bibitem[Goodwin et~al.(2015)Goodwin, Gurtowski, Ethe-Sayers, Deshpande,
  Schatz, and McCombie]{nanocorr}
S.~Goodwin, J.~Gurtowski, S.~Ethe-Sayers, P.~Deshpande, M.~Schatz, and
  W.~McCombie.
\newblock Oxford {N}anopore sequencing, hybrid error correction, and de novo
  assembly of a eukaryotic genome.
\newblock \emph{Genome Res.}, 2015.

\bibitem[Huang et~al.(1990)Huang, Ariki, and Jack]{HuaAriJac90}
X.~Huang, Y.~Ariki, and M.~Jack.
\newblock \emph{{Hidden {Markov} Models for Speech Recognition}}.
\newblock Edinburgh University Press, Edinburgh, 1990.

\bibitem[Kingma and Ba(2015)]{KinBa15}
D.~Kingma and J.~Ba.
\newblock Adam: A method for stochastic optimization.
\newblock In \emph{International Conference on Learning Representations
  (ICLR)}, 2015.

\bibitem[Krogh et~al.(1994)Krogh, Brown, Mian, Sj{\"o}lander, and
  Haussler]{KroBroMiaSjoetal94}
A.~Krogh, M.~Brown, I.~S. Mian, K.~Sj{\"o}lander, and D.~Haussler.
\newblock Hidden {Markov} models in computational biology: Applications to
  protein modeling.
\newblock \emph{Journal of Molecular Biology}, 235:\penalty0 1501--1531, Feb.
  1994.

\bibitem[Laehnemann et~al.(2016)Laehnemann, Borkhardt, and
  McHardy]{LaeBorMcH16}
D.~Laehnemann, A.~Borkhardt, and A.~McHardy.
\newblock Denoising {DNA} deep sequencing data-- high-throughput sequencing
  errors and their corrections.
\newblock \emph{Brief Bioinform}, 17\penalty0 (1):\penalty0 154--179, 2016.

\bibitem[Lee et~al.(2017)Lee, Moon, Yoon, and Weissman]{LeeMooYooWei16}
B.~Lee, T.~Moon, S.~Yoon, and T.~Weissman.
\newblock {DUDE-S}eq: {F}ast, flexible, and robust denoising for targeted
  amplicon sequencing.
\newblock \emph{{PLoS ONE}, 12(7):e0181463}, 2017.

\bibitem[Moon et~al.(2016)Moon, Min, Lee, and Yoon]{MooMinLeeYoo16}
T.~Moon, S.~Min, B.~Lee, and S.~Yoon.
\newblock Neural universal discrete denosier.
\newblock In \emph{Neural Information Processing Systems (NIPS)}, 2016.

\bibitem[Motta et~al.(2011)Motta, Ordentlich, Ramirez, Seroussi, and
  Weinberger]{MotOrdRamSerWei11}
G.~Motta, E.~Ordentlich, I.~Ramirez, G.~Seroussi, and M.~J. Weinberger.
\newblock The i{DUDE} framework for grayscale image denoising.
\newblock \emph{IEEE Trans. Image Processing}, 20:\penalty0 1--21, 2011.

\bibitem[Ordentlich et~al.(2003)Ordentlich, Seroussi, Verd{\'u}, Weinberger,
  and Weissman]{Ordetal03}
E.~Ordentlich, G.~Seroussi, S.~Verd{\'u}, M.~Weinberger, and T.~Weissman.
\newblock A universal discrete image denoiser and its application to binary
  images.
\newblock In \emph{IEEE ICIP}, 2003.

\bibitem[Ordentlich et~al.(2008)Ordentlich, Seroussi, Verd{\'u}, and
  Viswanathan]{OrdSerVerVis08}
E.~Ordentlich, G.~Seroussi, S.~Verd{\'u}, and K.~Viswanathan.
\newblock Universal algorithms for channel decoding of uncompressed sources.
\newblock \emph{IEEE Trans. Inform. Theory}, 54\penalty0 (5):\penalty0
  2243--2262, 2008.

\bibitem[Romero et~al.(2017)Romero, Kim, Giannakis, and Lopez-Valcarce]{Rom17}
D.~Romero, S.-J. Kim, G.~B. Giannakis, and R.~Lopez-Valcarce.
\newblock Learning power spectrum maps from quantized power measurements.
\newblock \emph{{IEEE} Trans. Signal Process.}, 2547 - 2560, 2017.

\bibitem[Ryu and Kim(2018)]{RyuKim18}
J.~Ryu and Y.-H. Kim.
\newblock Conditional distribution learning with neural networks and its
  application to universal image denoising.
\newblock pages 3214--3218, 10 2018.
\newblock \doi{10.1109/ICIP.2018.8451573}.

\bibitem[Weissman et~al.(2005)Weissman, Ordentlich, Seroussi, Verdu, and
  Weinberger]{Dude}
T.~Weissman, E.~Ordentlich, G.~Seroussi, S.~Verdu, and M.~Weinberger.
\newblock Universal discrete denoising: {K}nown channel.
\newblock \emph{{IEEE} Trans. Inform. Theory}, 51\penalty0 (1):\penalty0 5--28,
  2005.

\bibitem[Weissman et~al.(2007)Weissman, Ordentlich, Weinberger, Somekh-Baruch,
  and Merhav]{UFP06}
T.~Weissman, E.~Ordentlich, M.~Weinberger, A.~Somekh-Baruch, and N.~Merhav.
\newblock Universal filtering via prediction.
\newblock \emph{{IEEE} Trans. Inform. Theory}, 53\penalty0 (4):\penalty0
  1253--1264, 2007.

\end{thebibliography}

\section{Derivation of the M-step of ICE}

% \subsection{Approximate E-step}
% \begin{align}
% &\log p(Z^n;\Pib^{(t)})
% \geq \sum_{x^n} Q(x^n)\log\frac{p(x^n,Z^n;\Pib^{(t)})}{\tilde{Q}(x^n;\Pib^{(t)})}\nonumber\\
% =&\log p(Z^n;\Pib^{(t)})-D_{KL}\Big(Q(x^n)\|p(x^n|Z^n;\Pib^{(t)})\Big)\label{eq:lower bound}
% \end{align}

% Using variational posterior, we set

% \begin{align}
% Q(x^n)=\tilde{Q}(x^n;\Pib^{(t)})\triangleq&\prod_{i=1}^nq(x_i|Z_{i-k}^{i+k};\wb^{(t+1)})\label{eq:Approx E-step}
% \end{align}
% where,
% \begin{align}
% q(x_i=j|Z_{i-k}^{i+k};\wb^{(t+1)})=\sum_{s:s(Z_i)=j}\mathbf{p}^k(\wb^{(t+1)},\Cb_i)_s\label{eq:ind_post}
% \end{align}

% \subsection{M-step using Approximate E-step}
Consider the maximization problem of (Eq.(11), Manuscript) using $Q(x^n)$ as in (Eq.(12), Manuscript) to obtain the updated $\Pib^{(t+1)}$ from $\Pib^{(t)}$.
\begin{align}
\Pib^{(t+1)}
=&\underset{\mathbf{\Pi^{(t)}}}{\arg\max}\sum_{x^n}\tilde{Q}(x^n;\Pib^{(t)})\log \frac{p(x^n,Z^n;\Pib^{(t)})}{\tilde{Q}(x^n;\Pib^{(t)})}\nonumber\\
=&\underset{\mathbf{\Pi^{(t)}}}{\arg\max}\sum_{x^n}\tilde{Q}(x^n;\Pib^{(t)})\log p(x^n,Z^n;\Pib^{(t)})\nonumber\\
=&\underset{\mathbf{\Pi^{(t)}}}{\arg\max}\sum_{x^n}\tilde{Q}(x^n;\Pib^{(t)})\log p_{\mcX}(x^n)p(Z^n|x^n;\Pib^{(t)})\nonumber\\
=&\underset{\mathbf{\Pi^{(t)}}}{\arg\max}\sum_{x^n}\tilde{Q}(x^n;\Pib^{(t)})\log \prod_{i=1}^n\Pib^{(t)}(x_i,Z_i)\nonumber\\
=&\underset{\mathbf{\Pi^{(t)}}}{\arg\max}\sum_{x^n}\tilde{Q}(x^n;\Pib^{(t)})\sum_{i=1}^{n}\sum_{j=1}^{|\mcX|}\sum_{k=1}^{|\mcZ|}\mathds{1}_{\{x_i=j,Z_i=k\}}\log \Pib^{(t)}(j,k)\label{eq:obj}
\end{align}

Since above maximization process has following constraint, $\sum_{k=1}^{|\mcZ|}\Pib^{(t)}_{jk}=1$, we can consider Lagrangian of (\ref{eq:obj})

\begin{align}
L(\Pib^{(t)}, \lambdab)
=&\sum_{x^n}\tilde{Q}(x^n;\Pib^{(t)})\sum_{i=1}^{n}\sum_{j=1}^{|\mcX|}\sum_{k=1}^{|\mcZ|}\mathds{1}_{\{x_i=j,Z_i=k\}}\log \Pib^{(t)}(j,k)+\sum_{j=1}^{|\mcX|}\lambdab_j(1-\sum_{k=1}^{|\mcZ|}\Pib^{(t)}(j,k))\label{eq:lower}
\end{align}

Then, to apply the KKT conditon, the partial derivatives of the  Lagrangian  w.r.t $\Pib^{(t)}(j,k)$ and $\lambdab_j$ becomes

\begin{align}
\frac{\partial L}{\partial \Pib^{(t)}(j,k)}
=&\frac{1}{\Pib^{(t)}(j,k)}\sum_{x^n}\tilde{Q}(x^n;\Pib^{(t)})\sum_{i=1}^{n}\mathds{1}_{\{x_i=j,Z_i=k\}}-\lambdab_j\nonumber\\
\frac{\partial L}{\partial \lambdab_j}=&1-\sum_{k=1}^{|\mcZ|}\Pib^{(t)}_{jk}. \label{eq: lambda derivative}
\end{align}

Now, assume that the parameter satisfying KKT condition is $\Pib^{(t+1)}$. Then,

\begin{align}
\Pib^{(t+1)}(j,k)=\frac{1}{\lambdab_j}\sum_{x^n}\tilde{Q}(x^n;\Pib^{(t)})\sum_{i=1}^{n}\mathds{1}_{\{x_i=j,Z_i=k\}}\label{eq:Pi with lambda}
\end{align}

Using (\ref{eq: lambda derivative}) and (\ref{eq:Pi with lambda}),

\begin{align}
\lambdab_j=&\sum_{x^n}\tilde{Q}(x^n;\Pib^{(t)})\sum_{i=1}^{n}\mathds{1}_{\{x_i=j\}}\nonumber\\
\therefore
\Pib^{(t+1)}(j,k) =& \frac{\sum_{x^n}\tilde{Q}(x^n;\Pib^{(t)})\sum_{i=1}^{n}\mathds{1}_{\{x_i=j,Z_i=k\}}}{\sum_{x^n}\tilde{Q}(x^n;\Pib^{(t)})\sum_{i=1}^{n}\mathds{1}_{\{x_i=j\}}}\label{eq:semi-M-step}
\end{align}

However, we can change (\ref{eq:semi-M-step}) to an expression that contains (Eq.(12), Manuscript).

\begin{align}
\sum_{x^n}\tilde{Q}(x^n;\Pib^{(t)})\sum_{i=1}^{n}\mathds{1}_{\{x_i=j,Z_i=k\}}
=&\sum_{i=1}^{n}\sum_{x^n}\mathds{1}_{\{x_i=j,Z_i=k\}}\tilde{Q}(x^n;\Pib^{(t)})\nonumber\\
=&\sum_{i=1}^{n}\sum_{x^n}\mathds{1}_{\{x_i=j,Z_i=k\}}\prod_{l=1}^nq(x_l|Z_{l-k}^{l+k};\wb^{(t+1)})\nonumber\\
=&\sum_{i=1}^{n}\sum_{x_i}\mathds{1}_{\{x_i=j,Z_i=k\}}q(x_i|Z_{i-k}^{i+k};\wb^{(t+1)})\nonumber\\
=&\sum_{i=1}^{n}\mathds{1}_{\{Z_i=k\}}q(x_i=j|Z_{i-k}^{i+k};\wb^{(t+1)})\nonumber
\end{align}

Likewise,

\begin{align}
\sum_{x^n}\tilde{Q}(x^n;\Pib^{(t)})\sum_{i=1}^{n}\mathds{1}_{\{x_i=j\}}
=
\sum_{i=1}^{n}q(x_i=j|Z_{i-k}^{i+k};\wb^{(t+1)})\nonumber\\
\therefore
\Pib^{(t+1)}(j,k)=\frac{\sum_{i=1}^{n}\mathds{1}_{\{Z_i=k\}}q(x_i=j|Z_{i-k}^{i+k};\wb^{(t+1)})}{\sum_{i=1}^{n}q(x_i=j|Z_{i-k}^{i+k};\wb^{(t+1)})}\nonumber
\end{align}

% \newpage

\section{Experimental Details}
\subsection{Output dimension reduction}

\begin{figure}[ht]
    \centering
    \includegraphics[width=0.5\linewidth]{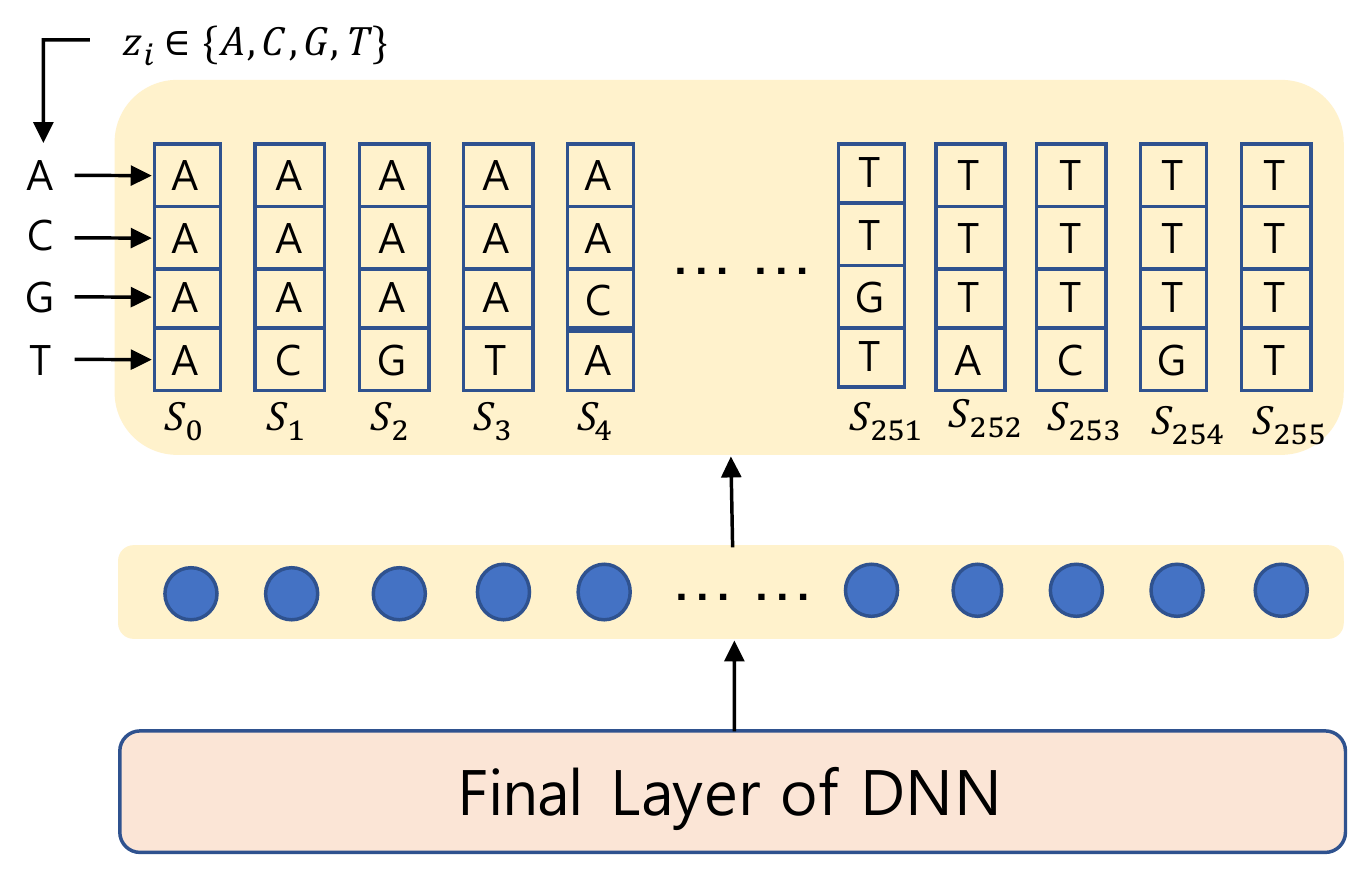}
    \caption{The original output layer of N-DUDE for DNA data. }\label{fig:256mapping}
\end{figure}

As a separate contribution, we also address one additional limitation of N-DUDE. Namely, the original N-DUDE has output size of $|\mathcal{S}|=|\mathcal{Z}|^{|\hat{\mathcal{X}}|}$, which can quickly grow very large when the alphabet size of the data grows. For example, even for DNA sequence that has alphabet size of 4, the output size of $\mathbf{p}^k(\wb,\cdot)$ becomes $|\mcS|=4^4=256$ as shown in Figure \ref{fig:256mapping}.
Such exponential growth of the output size may cause overfitting and inaccurate approximation for the induced posterioir (Eq.(8), Manuscript). 

% Before devising a channel estimation algorithm, we first try to solve the problem of exponentially large mappings in original Neural DUDE. Because of massive size of mappings, the $\Ellb_{\text{new}}$ matrix whose dimension is determined by $|S|$ becomes so uniform that the output distribution of classifier becomes uniform. Also, it is not scalable to other applications which has large alphabet size. Therefore, we try to minimize the number of mappings reasonably to get scalable and well performing model. 

\begin{figure}[ht]
    \centering
    \subfigure[16 mappings]{\label{fig:16map}
    \includegraphics[width=0.6\columnwidth]{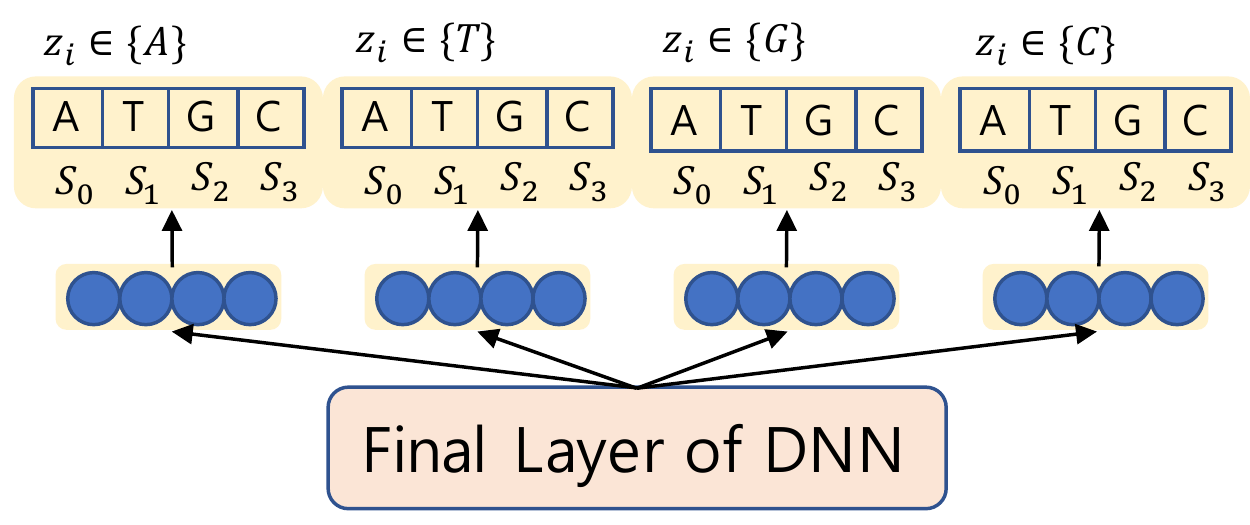}}
    \subfigure[5 mappings]{\label{fig:5map}
    \includegraphics[width=0.3\columnwidth]{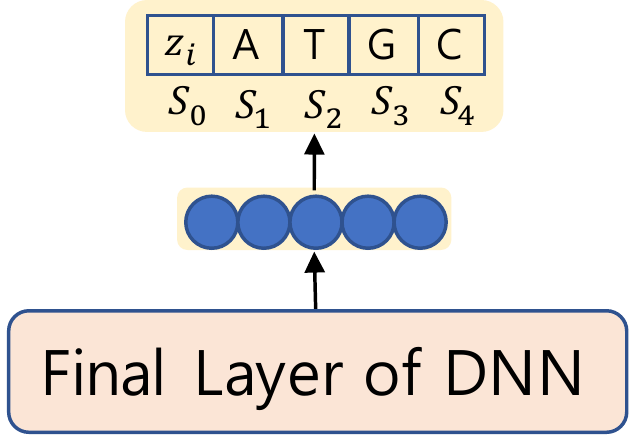}}
    \caption{The reduced output layer of N-DUDE for DNA data. }\label{fig:dim_reduction}
\end{figure}

In order to make ICE-N-DUDE more scalable with large alphabet size, we considered two output dimension reduction methods as shown in Figure \ref{fig:dim_reduction}, shown with the DNA example. First, Figure \ref{fig:16map} shows reducing the output size to $|\mcXhat||\mcZ|=16$ by implementing $|\mcZ|$ different output layers having $|\mcX|$ outputs. Note all 256 mappings in Figure \ref{fig:256mapping} can be enumerated by combining the \emph{partial} mappings for each $Z_i$ given in Figure \ref{fig:16map}. Second, Figure \ref{fig:5map} shows further reducing the output size to $|\mcXhat|+1=5$. That is, by simplifying the denoising to either ``saying-what-you-see'' (i.e., $s(Z_i)=Z_i$) or ``saying-one-in-$\mcXhat$, we can work with this reduced output size. With this reduction, the unnecessary variance in the model could reduce and the summation in (Eq.(8), Manuscript) would always involve only two mappings, hence, the approximation quality of $D_{KL}(\tilde{Q}(x^n;\Pib^{(t)})\|p(x^n|Z^n;\Pib^{(t)}))$ could improve. In fact, in our experimental results, we observed the second reduction yields much better denoising as well as the channel estimation results, hence, all of our results regarding N-DUDE employ the second reduction structure. 

\subsection{Training details for binary image denoising}
% \& DNA denoising}
In the binary image denoising experiment (Section 4.2), we used the first 10 images in PASCAL and all 8 images for Standard for estimating the channel before carrying out the denoising in each set. For ICE-N-DUDE and ICE-CUDE, the estimated channel by ICE was plugged-in, and the network parameters were fine-tuned for each image \emph{separately}. For fair comparison, N-DUDE($\Pib$) and CUDE($\Pib$)  were also first trained with the same images as ICE and BW, before fine-tuning for each image.
% \newpage
\subsection{True channels}
We used three different asymmetric $\Pib$'s in binary image denoising experiments with each $\Pib$ having average noise level of 0.1, 0.2 and 0.3.

% Three different $\Pib$s are as below.

\begin{align}
\Pib_{0.1}=\begin{bmatrix} 0.88 & 0.12 \\ 0.09 & 0.91 \end{bmatrix}\nonumber\\
\Pib_{0.2}=\begin{bmatrix} 0.83 & 0.17 \\ 0.23 & 0.77 \end{bmatrix}\nonumber\\
\Pib_{0.3}=\begin{bmatrix} 0.72 & 0.28 \\ 0.33 & 0.67 \end{bmatrix}\nonumber
\end{align}

For the DNA experiment, we used $4\times4$ asymmetric $\Pib$ as below. 

% $\Pib_{DNA}$ is as below.
\begin{align}
\Pib_{DNA}=\begin{bmatrix} 0.8122 & 0.0034 & 0.0894 & 0.0950 \\ 0.0096 & 0.8237 & 0.0808 & 0.0859 \\ 0.1066 & 0.0436 & 0.7774 & 0.0724 \\ 0.0704 & 0.0690 & 0.0889 & 0.7717\\ \end{bmatrix}\nonumber
\end{align}

\end{document}